\documentclass[sigconf,nonacm]{acmart}

\usepackage{amsmath,amssymb,amsfonts}
\usepackage{graphicx}
\usepackage{textcomp}
\usepackage{xcolor}
\usepackage{bm}
\usepackage{amsthm}
\usepackage{algorithmic}
\usepackage{amsmath}
\usepackage{hyperref}
\usepackage{subcaption}
\usepackage{xcolor}
\usepackage{algorithm}
\newtheorem{definition}{Definition}
\usepackage[T1]{fontenc}    
\AtBeginDocument{%
  }

\usepackage{microtype} 

\begin{document}

\title{Pareto Optimization with Robust Evaluation\\for Noisy Subset Selection}

\author{Yiheng Xu}
\affiliation{%
  \institution{State Key Laboratory for Novel Software Technology}
  \institution{School of Artificial Intelligence}
  \institution{Nanjing University}
  \city{Nanjing}
  \country{China}}
\email{xuyh@lamda.nju.edu.cn}

\author{Danxuan Liu}
\affiliation{%
  \institution{State Key Laboratory for Novel Software Technology}
  \institution{School of Artificial Intelligence}
  \institution{Nanjing University}
  \city{Nanjing}
  \country{China}}
\email{liudx@lamda.nju.edu.cn}

\author{Bin Zhang}
\affiliation{%
  \institution{State Key Laboratory of Technology and Equipment for Defense against Power System Operational Risks}
  \institution{Nari Technology Co., Ltd.}
  \city{Nanjing}
  \country{China}}
\email{zhangbin@sgepri.sgcc.com.cn}

\author{Weiyong Yang}
\affiliation{%
  \institution{State Key Laboratory of Technology and Equipment for Defense against Power System Operational Risks}
  \institution{Nari Technology Co., Ltd.}
  \city{Nanjing}
  \country{China}}
\email{yangweiyong@sgepri.sgcc.com.cn}

\author{Chao Qian}
\affiliation{%
  \institution{State Key Laboratory for Novel Software Technology}
  \institution{School of Artificial Intelligence}
  \institution{Nanjing University}
  \city{Nanjing}
  \country{China}}
\email{qianc@nju.edu.cn}
\begin{abstract}
  Subset selection is a fundamental problem in combinatorial optimization, which has a wide range of applications such as influence maximization and sparse regression. The goal is to select a subset of limited size from a ground set in order to maximize a given objective function. However, the evaluation of the objective function in real-world scenarios is often noisy. Previous algorithms, including the greedy algorithm and multi-objective evolutionary algorithms POSS and PONSS, either struggle in noisy environments or consume excessive computational resources. In this paper, we focus on the noisy subset selection problem with a cardinality constraint, where the evaluation of a subset is noisy. We propose a novel approach based on Pareto Optimization with Robust Evaluation for noisy subset selection (PORE), which maximizes a robust evaluation function and minimizes the subset size simultaneously. PORE can efficiently identify well-structured solutions and handle computational resources, addressing the limitations observed in PONSS. Our experiments, conducted on real-world datasets for influence maximization and sparse regression, demonstrate that PORE significantly outperforms previous methods, including the classical greedy algorithm, POSS, and PONSS. Further validation through ablation studies confirms the effectiveness of our robust evaluation function.
\end{abstract}



\maketitle

\section{Introduction}
The subset selection problem is a fundamental challenge in combinatorial optimization, aiming to select a subset $S$ with limited size $k$ from a ground set $V$ to maximize an objective function $f$, i.e. $\mathop{\arg\max}_{S \subseteq V}  f(S)$ \text{s.t.} $|S| \leq k,$ where $f: 2^V \rightarrow \mathbb{R}$ is monotone submodular. Since submodularity is an attractive property that encodes natural diminishing returns. This problem is generally NP-hard and is prevalent in numerous applications, like influence maximization~\cite{kempe2003maximizing, chen2009efficient}, sparse regression~\cite{miller2002subset, das2011submodular}, feature selection~\cite{hall1998practical, feng2019feature}, maximum coverage~\cite{feige1998threshold, hochbaum1998analysis}, human assisted learning~\cite{liu2023human}, peptide vaccine design~\cite{liu2024peptide} and migrant resettlement~\cite{liu2024migrant} etc.

It is known that the classical greedy algorithm, which iteratively chooses an item with the largest marginal gain on $f$, can achieve an optimal polynomial-time approximation guarantee of $1-1/e$~\cite{nemhauser1978best}. However, the performance of the greedy algorithm may be limited by its greedy nature. A Pareto Optimization approach for Subset Selection, namely POSS~\cite{qian2015subset, qian2020subset}, reformulates the subset selection problem as a bi-objective optimization problem. POSS maximizes the objective $f$ and minimizes the subset size simultaneously, employing multi-objective evolutionary algorithms to solve it. POSS also achieves the optimal approximation guarantee of $1-1/e$ and has demonstrated superior performance on real-world tasks. While research on POSS and its various variants has been well-established~\cite{qian2021multiobjective, qian2023multi, qian2022result, qian2017subsetcost, qian2019maximizing, qian2018multiset, qian2017optimizing, liu2023result, gu2023subset, qian2023can, liu2024biased, roostapour2022pareto, yan2024sliding, bian2022robust, neumann2024sliding, qian2018distributed, bian2021fast, qian2017constrained, qian2018sequence, yan2024sampling, neumann2020optimising, do2020maximizing,  neumann2023fast, neumann2023benchmarking}, it typically relies on the assumption that the objective function can be evaluated accurately.

In many real-world scenarios, however, we can only obtain a noisy value of the objective function rather than an accurate evaluation. For example, for influence maximization, calculating the accurate influence spread objective is challenging~\cite{chen2010scalable}. The approximate value is often used, calculated by simulating random diffusion, which introduces noise. Assuming that the noisy function value $F(S)$ is a random variable (i.e., random noise) and the expectation of $F(S)$ corresponds to the accurate value $f(S)$, it has been proven that the greedy algorithms using sampling methods to assist evaluation can achieve nearly a $(1-1/e)$ approximation guarantee~\cite{kempe2003maximizing, singla2016noisy}. For more works related to noisy subset selection with a monotone and submodular objective function, we refer to~\cite{hassidim2017submodular,hassidim2018optimization}.

Qian \textit{et al.}~\cite{qian2017subset} considered a more general noisy subset selection problem, extending the objective function $f$ from the submodular case to the general case where $f$ is not necessarily submodular. They proved that the classical greedy algorithm, which iteratively chooses an item with the largest marginal gain on $F$, achieves an $\frac{(1-\epsilon)\gamma_{\hat{S},k}}{2\epsilon k + (1-\epsilon)\gamma_{\hat{S},k}}(1-(\frac{1-\epsilon}{1+\epsilon})^k(1-\frac{\gamma_{\hat{S},k}}{k})^k)$-approximation guarantee, where $\hat{S}$ denotes the subset output by the greedy algorithm, $\gamma_{\hat{S},k}$ characterizes how close a set function $f$ is to submodularity, and $\epsilon$ quantifies the extent to which the approximation $F(S)$ can differ from the actual value $f(S)$, i.e., $(1-\epsilon)f(S)\leq F(S) \leq (1+\epsilon)f(S)$. They also proved that POSS can achieve nearly the same approximation ratio as that of the classical greedy algorithm. As the classical greedy algorithm and POSS are both designed for noise-free objective functions, a modified method called PONSS was further proposed~\cite{qian2017subset}. This method incorporates a noise-aware comparison strategy into POSS, adapting it to perform effectively even when the evaluation of the objective function is noisy. When comparing two solutions that have closely noisy objective values, POSS selects the solution with a better observed value. In contrast, PONSS retains both solutions unless the ratio between them exceeds a specified threshold, thereby reducing the risk of discarding a potentially good solution. It is proved that PONSS can obtain a $\frac{1-\epsilon}{1+\epsilon}(1-e^{-\gamma})$-approximation guarantee, where $\gamma=\min_{S:|S|=k-1}\gamma_{S,k}$. Particularly for the submodular case where $\gamma = 1$, PONSS can achieve a constant approximation ratio even when $\epsilon$ is a constant, while the greedy algorithm and POSS only guarantee a $\Theta(1/k)$ approximation ratio. PONSS exhibits superior performance over both POSS and the classical greedy algorithm in real-world applications. Additionally, to address the challenges of large-scale noisy subset selection, Qian~\cite{Qian20} proposed a distributed version of PONSS.

While PONSS effectively mitigates noise via its noise-aware comparison strategy, it incurs substantial computational overhead by re-evaluating solutions to prevent high-quality candidates from being inadvertently discarded. Specifically, PONSS may require $2B$ additional evaluations each time the population is updated, where $B$ is a hyperparameter. To achieve the theoretical guarantee mentioned above, PONSS requires $2eBnk^2\log 2k$ evaluations, which is $B\log 2k$ times that of POSS.

In this paper, we propose a new method based on Pareto Optimization with Robust Evaluation for noisy subset selection, briefly called PORE. Unlike POSS and PONSS, PORE maximizes a robust evaluation function and minimizes the size of subset simultaneously, and then employs multi-objective evolutionary algorithms to solve it. To obtain the robust evaluation function value of a solution $S$, one needs to calculate the average value of the noisy function $F$ over all subsets of $S$ that have a size equal to $|S|-1$. This can better identify the well-structured solutions, using much less evaluations compared to the strategy of re-evaluating solutions during the population update of PONSS. In addition, PORE utilizes the advantage of PONSS by inheriting its noise-aware comparison strategy. We conducted experiments on influence maximization and sparse regression problems. The results on real-world datasets under various noise levels show that PORE clearly outperforms previous algorithms, namely, the classical greedy algorithm, POSS, and PONSS. In most settings, PORE has achieved a performance improvement of over 5\% compared to current state-of-the-art algorithms, and in the \textit{protein} dataset for sparse regression problem, it has consistently achieved more than a 20\% performance improvement in each setting. Additionally, the PORE algorithm has shown greater stability compared to these benchmarks. The ablation studies show the effectiveness of the robust evaluation function. 

The remainder of this paper begins with an introduction to the preliminaries. We then explore in detail the greedy, POSS, and PONSS algorithms. Next, we introduce our proposed algorithm, PORE, and demonstrate its efficacy through experimental results. Specifically, the PORE algorithm significantly outperforms previous methods in two distinct applications: influence maximization and sparse regression. We further substantiate the sound design of the algorithm through ablation studies and investigate how noise intensity and the hyperparameter $\theta$ influence its performance. The paper concludes by summarizing our findings.

\section{Noisy Subset Selection}

Given a finite non-empty set $V=\{v_1,v_2,\ldots,v_n\}$, we study the functions $f: 2^V \rightarrow \mathbb{R} $ defined on subsets of $V$, where $\mathbb{R}$ denotes the set of reals. We study the noisy subset selection problem with a cardinality constraint, as presented in Definition~\ref{def1}, aiming to select a subset $S$ from the finite set $V$ that maximizes the given objective $f$, while ensuring that the subset size $|S|$ does not exceed $k$.

\begin{definition}[Noisy Subset Selection with a Cardinality Constraint]\label{def1}
    Given a ground set $V=\{v_1, v_2, \dots, v_n\}$, a monotone objective function $f: 2^V \rightarrow \mathbb{R} $  and an integer $k$, to find a subset $S$ of at most $k$ items maximizing $f$, i.e.,
    \begin{equation} \label{eq1}
        {\arg \max}_{S\subseteq V} f(S) \quad  s.t. \quad |S| \leq k.
    \end{equation}
However, only a noisy function $F(S)$ can be calculated, where $F(S)$ is a random variable (i.e., the objective evaluation is subject to random noise) and the expectation of $F(S)$ corresponds to the accurate value $f(S)$.
\end{definition}

A set function $f$ is monotone if for any $A\subseteq B\subseteq V$, $f(A) \leq f(B)$. We assume w.l.o.g. that the monotone function $f$ is normalized, i.e., $f(\emptyset)=0$. Note that the objective function $f$ in Eq.~(\ref{eq1}) does not need to hold the diminishing returns property, that is, it does not need to satisfy $\forall A \subseteq B\subseteq V, v \notin B: f(A \cup \{v\})-f(A) \geq f(B \cup \{v\}) - f(B).$ The submodularity ratio in Definition~\ref{def4} characterizes how close a set function $f$ is to submodularity. It is easy to see that $f$ is submodular iff $\gamma_{S,k}(f) = 1$ for any $S$ and $k$.  When $f$ is clear, we will use $\gamma_{S,k}$ shortly.

\begin{definition}[Submodularity Ratio~\cite{das2011submodular, qian2018approximation}]\label{def4}
    Let $f$ be a non-ne\-gative set function. The submodularity ratio of $f$ with respect to a set $S$ and a parameter $k\geq 1$ is
    \begin{equation*}
        \gamma_{S, k}(f) = \min_{L\subseteq S, R:|R|\leq k, R\cap L = \emptyset} \frac{\sum_{v \in R}(f(L\cup \{v\})-f(L))}{f(L\cup R)-f(L)}. 
    \end{equation*}
\end{definition}
Now we introduce two applications that will be empirically studied in this paper, i.e., influence maximization and sparse regression.

Influence maximization~\cite{kempe2003maximizing} is to identify a subset of exceptionally influential users within a social network. As presented in Definition~\ref{IM}, it is formally presented as selecting a subset $S$ of up to $k$ nodes from a directed graph $G(V, E)$, to maximize the expected number of activated nodes, or influence spread, by propagating from this initial set $S$. Each edge $(u, v)\in E$ has an associated probability $p_{u,v}$ indicating the likelihood of influence transmission. Using the Independence Cascade (IC) model~\cite{goldenberg2001talk}, the influence dynamics begin with the initial subset $S$ and propagate through probabilistic interactions until no further activations occur. The set of nodes activated by propagating from $S$ is denoted as $IC(S)$, which is a random variable. The objective $\mathbb{E} [|IC(S)|]$ is monotone and submodular\cite{kempe2003maximizing}. Computing the influence spread for influence maximization is \#P-hard, as demonstrated by Chen et al.~\cite{chen2010scalable}. As a result, it is commonly estimated through simulations of the random diffusion process~\cite{kempe2003maximizing}, which introduces noise.

\begin{definition}[Influence Maximization~\cite{kempe2003maximizing}]\label{IM}
    Given a directed graph $G(V, E)$, edge probabilities $p_{u, v}$  $((u, v) \in E)$, and an integer $k$, the task is as follows:
    \begin{equation*}
        \arg \max\nolimits_{S \subseteq V} \mathbb{E} [|IC(S)|] \quad s.t. \quad |S|\leq k.
    \end{equation*}
\end{definition}

Sparse regression~\cite{miller2002subset} is to find a sparse approximation solution to the linear regression problem. As presented in Definition~\ref{SR}, given a set of observational variables $V=\{v_1, v_2, \dots, v_n\}$, a target predictor variable $z$ and an integer $k$, the goal of sparse regression is to select at most $k$ variables that minimize the \textit{mean squared error} $\text{MSE}_{z, S}$. An equivalent objective of sparse regression is often used, which is
\begin{equation*}
    \arg \max\nolimits_{S \subseteq V} R^2_{z, S} \quad s.t. \quad |S|\leq k,
\end{equation*}
where $R^2_{z, S} = (Var(z)-MSE_{z, S})/Var(z)$ is the \textit{squared multiple correlation}~\cite{diekhoff1992statistics, johnson2002applied}, and can be simplified to be $1-MSE_{z, S}$, because $z$ is assumed to be normalized to have variance 1 (i.e.,$Var(z) = 1$). It's important to note that the objective function $R_{z, S}^2$ is monotone, but non-submodular~\cite{das2011submodular}. For sparse regression, only a set of limited data can be used for evaluation, which makes the evaluation noisy.

\begin{definition}[Sparse Regression~\cite{miller2002subset}]\label{SR}
    Given all observation variables  $V=\{v_1, v_2, \dots, v_n\}$, a predictor variable $z$ and an integer $k$, the mean squared error of a subset $S \subseteq V$ is defined as
    \begin{equation*}
        MSE_{z, S} = \min\nolimits_{\bm{\alpha} \in \mathbb{R}^{|S|}}\mathbb{E}[(z-\sum\nolimits_{i\in S} \alpha_i v_i)^2].
    \end{equation*}
    \textit{Sparse regression is to find a set of at most $k$ variables minimizing the mean squared error, i.e.,}
    \begin{equation*}
        \arg \min\nolimits_{S \subseteq V} MSE_{z, S} \quad s.t. \quad |S|\leq k.
    \end{equation*}
\end{definition}

\section{Previous Algorithms}
We now introduce previous algorithms for the noisy subset selection problem with a cardinality constraint, i.e., the classical greedy algorithm~\cite{nemhauser1978best}, POSS~\cite{qian2015subset} and PONSS~\cite{qian2017subset}.

\subsection{The Greedy Algorithm}
\begin{algorithm}[t]
    \raggedright 
    \caption{Greedy Algorithm~\cite{nemhauser1978best}}\label{alg:greedy}
    \textbf{Input}:  all items $V=\{v_1, ... ,v_2\}$, a noisy objective function $F$, and an integer $k$ \\
    \textbf{Output}: a subset of $V$ with $k$ items\\
    \textbf{Process}:
    \begin{algorithmic}[1]
        \STATE Let $i = 0$ and $S_i = \emptyset$.
        \WHILE{$i < k$}
        \STATE \quad Let $v^* = \arg \max_{v\in V \setminus S_i} F(S_i \cup \{v\})$.
        \STATE \quad Let $S_{i+1} = S_i \cup \{v^*\}$, and $i=i+1$.
        \ENDWHILE
        \STATE \textbf{return} $S_k$
    \end{algorithmic}
\end{algorithm}

The classical greedy algorithm~\cite{nemhauser1978best}, as delineated in Algorithm~\ref{alg:greedy}, iteratively adds one item that maximizes the marginal improvement on the objective function $F$, until the subset size reaches $k$. For noisy subset selection in Definition~\ref{def1}, Qian \textit{et al.}~\cite{qian2017subset} proved that the classical greedy algorithm can find a subset $S$ with size at most $k$ satisfying
\begin{equation}\label{greedy-bound}
    f(S)\!\geq\!\frac{(1-\epsilon)\gamma_{S,k}}{2\epsilon \! +\! (1\!-\!\epsilon)\gamma_{S,k}}\big(1-\big(\frac{1-\epsilon}{1+\epsilon}\big)^k\big(1-\frac{\gamma_{S,k}}{k}\big)^k\big)\cdot OPT,
\end{equation}
where the submodularity ratio $\gamma_{S,k}$ introduced in Definition~\ref{def4} characterizes how close a set function $f$ is to submodularity, $\epsilon$ quantifies the extent to which the approximation $F(S)$ can differ from the actual value $f(S)$, i.e., $(1-\epsilon)f(S)\leq F(S) \leq (1+\epsilon)f(S)$, and $OPT$ denotes the optimal function value of Eq.~(\ref{eq1}), i.e., $OPT=\max_{S\subseteq V, |S|\leq k} f(S)$.


\subsection{The POSS Algorithm}
\begin{algorithm}[t] 
    \raggedright 
    \caption{POSS Algorithm~\cite{qian2015subset}}\label{alg:POSS}
    \textbf{Input}:  all items $V=\{v_1, ... ,v_2\}$, a noisy objective function $F$, and an integer $k$ \\
    \textbf{Output}: a subset of $V$ with at most $k$ items\\
    \textbf{Process}:
    \begin{algorithmic}[1]
    \STATE Let $\bm{x} = \{0\}^n$, $P=\{\bm{x}\}$.
    \REPEAT
    \STATE Select $\bm{x}$ from $P$ uniformly at random.
    \STATE Generate $\bm{x'}$ by flipping each bit of $\bm{x}$ with probability $\frac{1}{n}$.
    \IF {$\nexists \bm{z} \in P$ such that $\bm{z}\succ \bm{x'}$}
    \STATE $P = (P \setminus \{\bm{z} \in P \,|\, \bm{x'}\succeq \bm{z}\}) \cup \{\bm{x'}\}$.
    \ENDIF
    \UNTIL{some criterion is met}
    \STATE \textbf{return} $\arg \max_{\bm{x} \in P, |\bm{x}|\leq k} F(\bm{x})$
    \end{algorithmic}
\end{algorithm}

Let a Boolean vector $\bm{x} \in \{0, 1\}^n$ be a subset $S$ of $V$, where $x_i = 1$ if $v_i \in S$ and $x_i = 0$ otherwise. For convenience, we will not distinguish $\bm{x} \in \{0, 1\}^n$ and its corresponding subset. The POSS algorithm~\cite{qian2015subset} reformulates the original problem Eq.~(\ref{eq1}) as a bi-objective maximization problem:
\begin{align}\label{eq-bi-objective}
\arg\max\nolimits_{\bm{x} \in \{0,1\}^n} (f_1(\bm{x}), f_2(\bm{x})),
\end{align}
where
\begin{align*}
f_1(\bm{x}) = \begin{cases}
	-\infty, &{|\bm{x}|\ge 2k} \\
	F(\bm{x}), &{\text{otherwise}}
\end{cases},\quad
f_{2}(\bm{x})=-|\bm{x}|.
\end{align*}

After constructing the bi-objective problem in Eq.~(\ref{eq-bi-objective}), POSS employs the process of multi-objective evolutionary algorithms to simultaneously optimize these two objective functions. As described in Algorithm~\ref{alg:POSS}, POSS starts with the zero solution $\{0\}^n$ (line~1), and repeatedly improves the quality of solutions in population $P$ (lines~2--8). In each iteration, a parent solution $\bm{x}$ is selected from the population by uniform selection (line~3), and an offspring solution $\bm{x'}$ is then generated from the parent solution using the bit-wise mutation operator (line~4). As the two objectives may be conflicting, the domination relationship in Definition~\ref{def3} is often used for comparing two solutions. If the newly generated individual $\bm{x'}$ is not dominated by any individual currently in the population (line~5), it will be added to the population. Subsequently, the population will be updated, and those individuals weakly dominated by $\bm{x'}$ will be deleted from population (line~6). After running some iterations, the individual with the largest $F$ value in the population $P$ satisfying the cardinality constraint is returned (line~9).

\begin{definition}[Domination]\label{def3}
    For two solutions $\bm{x}$ and $\bm{y}$:
    
    \begin{itemize}
        \item $\bm{x}$ weakly dominates $\bm{y}$ (denoted as $\bm{x}\succeq \bm{y}$) if $f_1(\bm{x}) \geq f_1(\bm{y}) \wedge f_2(\bm{x}) \geq f_2(\bm{y})$;
        \item  $\bm{x}$ dominates $\bm{y}$ (denoted as $\bm{x} \succ  \bm{y}$) if $\bm{x}\succeq \bm{y}$ and $f_1(\bm{x}) > f_1(\bm{y}) \vee f_2(\bm{x}) > f_2(\bm{y})$.
    \end{itemize}
\end{definition}
When the objective function $f$ is noise-free and submodular, POSS is proven to achieve a $(1-\frac{1}{e})$-approximation ratio~\cite{FriedrichN15}. In scenarios where $f$ may be non-submodular,  POSS still demonstrates the best-so-far bound of $1-e^{-\gamma_{\emptyset, k}}$~\cite{qian2015subset}. Additionally, in real-world tasks, the POSS algorithm outperforms the greedy algorithm~\cite{qian2015subset}. For noisy subset selection in Definition~\ref{def1}, Qian \textit{et al.}~\cite{qian2017subset} proved that POSS can find a subset $S$ with size at most $k$ satisfying
$$
    f(S) \geq \frac{(1-\epsilon)\gamma_{min}}{2\epsilon k + (1-\epsilon)\gamma_{min}}\big(1-\big(\frac{1-\epsilon}{1+\epsilon}\big)^k\big(1-\frac{\gamma_{min}}{k}\big)^k\big)\cdot OPT,
$$
where $\gamma_{\text{min}} \!=\! \min\limits_{S:|S|=k-1} \gamma_{S, k}$. This approximation ratio is nearly the same as that, i.e. Eq.~(\ref{greedy-bound}), of the classical greedy algorithm.

\subsection{The PONSS Algorithm}

The performance of the classical greedy algorithm and POSS may be unstable in the presence of noise, because they are designed for noise-free environment. Specifically, they compare and save the solution with the best observed value, even with noise, which may lead to a situation where a worse solution appears to have a better $F$ value and thus survives, replacing truly superior solutions. Qian \textit{et al.}~\cite{qian2017subset} then proposed an approach named PONSS, which was modified from POSS. Like POSS, PONSS reformulates the original problem Eq.~(\ref{eq1}) as a bi-objective maximization problem in Eq.~(\ref{eq-bi-objective}), and solves it following a multi-objective evolutionary process. Unlike POSS, PONSS employs a noise-aware comparison strategy to reduce the risk of discarding a good solution obscured by noise. That is, PONSS compares solutions based on $\theta$-domination as shown in Definition~\ref{def5}. Specifically, $\bm{x} \succeq_\theta \bm{y}$ if $f_1(\bm{x}) \geq \frac{1+\theta}{1-\theta} \cdot f_1(\bm{y})$ and $ |\bm{x}| \leq |\bm{y}|$.

\begin{definition}[$\theta$-Domination]\label{def5} 

For two solutions $\bm{x}$ and $\bm{y}$: 

\begin{itemize}
    \item $\bm{x}$ weakly $\theta$-dominates $\bm{y}$ (denoted as $\bm{x} \succeq_\theta \bm{y}$ ) if $f_1(\bm{x}) \geq \frac{1+\theta}{1-\theta} \cdot f_1(\bm{y})\wedge f_2(\bm{x}) \geq f_2(\bm{y})$;
    
    \item $\bm{x}$ $\theta$-dominates $\bm{y}$ (denoted as $\bm{x} \succ_{\theta} \bm{y}$) if $\bm{x} \succeq_\theta \bm{y}$ and $f_1(\bm{x}) > \frac{1+\theta}{1-\theta} \cdot f_1(\bm{y})\vee  f_2(\bm{x}) > f_2(\bm{y})$.
\end{itemize}

\end{definition}

The process of PONSS algorithm is shown in Algorithm~\ref{alg:PONSS}. PONSS compares solutions by $\theta$-domination and updates the population $P$ in lines~5--17. However, using $\theta$-domination will make the size of $P$ very large. PONSS then introduces a parameter $B$ to limit the number of individuals in the population $P$ for each possible subset size. If the number of individuals $\bm{z}\in P$ of size $|\bm{x}'|$ surpasses $B$ (lines~7--8), PONSS iteratively re-evaluates and compares two individuals randomly selected
from $Q$, and the superior one is retained (line~10--15). This process is repeated $B$ times to guarantee that no more than $B$ individuals of identical size remain in $P$.

\begin{algorithm}[t]
    \raggedright 
    \caption{PONSS Algorithm~\cite{qian2017subset}}\label{alg:PONSS}
    \textbf{Input}:  all items $V=\{v_1, ... ,v_2\}$, a noisy objective function $F$, and an integer $k$ \\
    \textbf{Parameter}: $\theta$ and $B$\\
    \textbf{Output}: a subset of $V$ with at most $k$ items\\
    \textbf{Process}:
    \begin{algorithmic}[1]
    \STATE Let $\bm{x} = \{0\}^n$, $P=\{\bm{x}\}$. 
    \REPEAT
    \STATE Select $\bm{x}$ from $P$ uniformly at random.
    \STATE Generate $\bm{x'}$ by flipping each bit of $\bm{x}$ with probability $\frac{1}{n}$.
    \IF {$\nexists \bm{z} \in P$ such that $\bm{z} \succ_{\theta} \bm{x'} $}
    \STATE $P=(P \setminus \{\bm{z} \in P \,|\,\bm{x'} \succeq_{\theta} \bm{z}\}) \cup \{\bm{x'}\}$.
    \STATE $Q=\{\bm{z} \in P \,| \,|\bm{z}|=|\bm{x'}|\}$.
    \IF {$|Q| = B+1$}
    \STATE $P = P\setminus Q$ and let $j = 0$.
    \WHILE{$j < B$}
    \STATE Select $\bm{z_1}$, $\bm{z_2}$ from $Q$ uniformly at random without replacement.
    \STATE Evaluate $F(\bm{z_1})$, $F(\bm{z_2})$; 
    \STATE Let $\widehat{\bm{z}} = \arg \max_{\bm{z} \in \{\bm{z_1}, \bm{z_2}\}} F(\bm{z})$ (breaking ties randomly).
    \STATE $P = P \cup \{\widehat{\bm{z}}\}$, $Q = Q\setminus \{\widehat{\bm{z}}\}$, and $j = j+1$.
    \ENDWHILE
    \ENDIF 
    \ENDIF
    \UNTIL{some criterion is met}
    \STATE \textbf{return} $\arg \max_{\bm{x} \in P, |\bm{x}|\leq k} F(\bm{x})$
    \end{algorithmic}
\end{algorithm}

For noisy subset selection in Definition~\ref{def1}, PONSS~\cite{qian2017subset} can find a subset $S$ with size at most $k$ 
satisfying
$$
     f(S) \geq \frac{1-\epsilon}{1+\epsilon}\big(1-\big(1-\frac{\gamma_{\text{min}}}{k}\big)^k\big)\cdot OPT.
$$
For the submodular case where $\gamma_{\text{min}}=1$, PONSS can achieve a constant approximation ratio even with a constant $\epsilon$, significantly outperforming the greedy algorithm and POSS, which only provide a $\Theta(1/k)$ approximation rate. PONSS retains solutions with close $F$ value through $\theta$-domination, reducing the risk of superior solutions being erroneously replaced by inferior ones. Consequently, in noisy real-world scenarios, it consistently demonstrates superior performance over both POSS and the greedy algorithm.

Although $\theta$-domination can prevent inferior solutions from eliminating superior ones, PONSS still encounters efficiency bottlenecks during the process of population update. To ensure robustness to noise and limit the population size, PONSS re-evaluates individuals during each population update, resulting in an additional computational cost of $2B$. To obtain a solution with theoretical guarantee, the computational cost of the PONSS algorithm is $B\log 2k$ times that of the POSS algorithm.

\section{The Proposed PORE algorithm}\label{section-PORE}
In this section, we propose a new method based on Pareto Optimization with Robust Evaluation for noisy subset selection, briefly called PORE, which assesses the quality of individuals using fewer evaluation resources than PONSS. 

PORE reformulates the original problem Eq.~(\ref{eq1}) as a bi-objective maximization problem:
\begin{align}\label{eq-bi-PORE}
\arg\max\nolimits_{\bm{x} \in \{0,1\}^n} (f_1(\bm{x}), f_2(\bm{x})),
\end{align}
where \begin{align*}
f_1(\bm{x})=\begin{cases} 
        -\infty ,  \quad &\bm{x} \geq 2k, \\
        \frac{\sum_{\bm{y}\subseteq \bm{x}, |\bm{y}| = |\bm{x}|-1 } F(\bm{y})}{|\bm{x}|} , &\text{otherwise} 
        \end{cases},\quad
f_{2}(\bm{x})=-|\bm{x}|.
\end{align*}
The function $f_1$ is computed by a robust evaluation, which calculates the average value of the noisy function $F$ over all subsets of $S$ with $|S|-1$ elements. This method provides a more nuanced understanding of the stability and reliability of the current solution by quantifying the impact of minor structural perturbations on its performance. Consequently, this approach is more effective in identifying well-structured or robust solutions, namely those that are less susceptible to disturbances caused by noise. Similar ideas are also used in~\cite{hassidim2017submodular,hassidim2018optimization,huang2022efficient}. Similar to POSS and PONSS, PORE calculates the first objective $f_1(\bm{x})$ for solutions smaller than $2k$, aiming to have some possibility of generating a desired solution through a parent solution with size from $k+1$ to $2k-1$.

After constructing the bi-objective problem in Eq.~(\ref{eq-bi-PORE}), PORE follows a process of multi-objective evolutionary algorithms to maximize the robust evaluation function and minimize the size of subset simultaneously, as shown in Algorithm~\ref{alg:PORE}. It still starts from the zero solution $\{0\}^n$ (line~1), and repeatedly improves the quality of solutions in the population $P$ (lines~2–13). In each iteration, it uniformly selects a parent solution $\bm{x}$ at random (line~3), and generates an offspring solution $\bm{x}'$ by bit-wise mutation (line~4). Then, the robust evaluation is used to evaluate $\bm{x}'$ (line~5). Like PONSS, PORE also uses $\theta$-domination to compare solutions. If no solution $\bm{z}\in P$ $\theta$-dominates the newly generated solution $\bm{x}'$, then the solutions that are weakly $\theta$-dominated by $\bm{x}'$ will be deleted from the population $P$ (lines 6--7). When the population contains more than $B$ individuals with size $|\bm{x}'|$, PORE will directly discard the individual with the lowest evaluation value $f_1$ (lines~9--11) without re-evaluation, which is different from PONSS. This is because the goodness of individuals is comparatively precise by robust evaluation. After running some iterations, the individual with the largest $f_1$ value in the population $P$ satisfying the cardinality constraint is returned (line~14). Note that the aim of PORE is to find a good solution of the original problem in Definition~\ref{def1}, rather than the Pareto front\footnote{A solution is Pareto optimal if no other solution dominates it. The collection of objective vectors of all Pareto optimal solutions is called the Pareto front.} of the reformulated bi-objective problem in Eq.~(\ref{eq-bi-PORE}). That is, the bi-objective reformulation is an intermediate process.

Note that, when the population contains more than $B$ individuals of a same size, PONSS randomly selects two individuals at a time for re-evaluation, and keeps the better one. This re-evaluation process will repeat $B$ times to provide truly superior solutions with greater chances to persist in the population, thus mitigating the effects of noise. This results in a total of $2B$ additional evaluations in each iteration. Compared to PONSS, PORE spends computational resources more rationally on the robust evaluations, ensuring that each evaluation is already quite accurate and eliminating the need for extensive computations during population updates.

\begin{algorithm}[t]
    \raggedright 
    \caption{PORE}\label{alg:PORE}
    \textbf{Input}:  all items $V=\{v_1, ... ,v_2\}$, a noisy objective function $F$, and an integer $k$ \\
    \textbf{Parameter}: $\theta$ and $B$\\
    \textbf{Output}: a subset of $V$ with at most $k$ items\\
    \textbf{Process}:
    \begin{algorithmic}[1]
    \STATE Let $\bm{x} = \{0\}^n$, $P=\{\bm{x}\}$. 
    \REPEAT
    \STATE Select $\bm{x}$ from $P$ uniformly at random.
    \STATE Generate $\bm{x'}$ by flipping each bit of $\bm{x}$ with probability $\frac{1}{n}$.
    \STATE Evaluate $\bm{x'}$ according to Eq.~(\ref{eq-bi-PORE}).
    
    \IF {$\nexists \bm{z} \in P$ such that $\bm{z} \succ_{\theta} \bm{x'} $}
    \STATE $P=(P\setminus \{\bm{z} \in P \,|\,\bm{x'} \succeq_{\theta} \bm{z}\}) \cup \{\bm{x'}\}$.
    \STATE $Q=\{\bm{z} \in P \,| \,|\bm{z}|=|\bm{x'}|\}$.
    \IF {$|Q| = B+1$}
    \STATE Eliminate the individual with the smallest $f_1$ value in Q from the population $P$.
    \ENDIF 
    \ENDIF
    \UNTIL{some criterion is met}
    \STATE \textbf{return} $\arg \max_{\bm{x} \in P, |\bm{x}|\leq k} f_1(\bm{x})$
    \end{algorithmic}
\end{algorithm}

\section{Empirical Study}
In this section, we compare the performance of PORE with the previous competitive algorithms PONSS and POSS, and set the greedy algorithm as a baseline. We conduct experiments on two applications: influence maximization and sparse regression. For randomized algorithms (POSS, PONSS and PORE), the same total number of evaluations is used for the fairness of comparison, i.e., $2ek^2n$, as set in~\cite{qian2017subset}. Note that, POSS requires only one evaluation of the objective function per iteration, while PONSS needs either $1$ or $1+2B$ evaluations depending on whether the condition in line~8 of Algorithm~\ref{alg:PONSS} is satisfied.  PORE requires $|\bm{x'}|$ evaluations in each iteration to estimate the fitness of the solution $\bm{x'}$ generated in line~4 of Algorithm~\ref{alg:PORE}. For both PONSS and PORE, the hyperparameter $B$ is set to $k$. Note that in each run of algorithm, only a noisy objective value $F$ can be obtained; while for the final solution returned by each algorithm, we report its accurately estimated $f$ value by an expensive evaluation. Given the inherent noise and randomness within these algorithms, we conduct 30 independent runs for each algorithm and report the average of the accurately estimated $f$ values and the standard deviation.

\begin{figure*}
    \begin{subfigure}{\textwidth}
      \centering
      \includegraphics[width=0.36\linewidth]{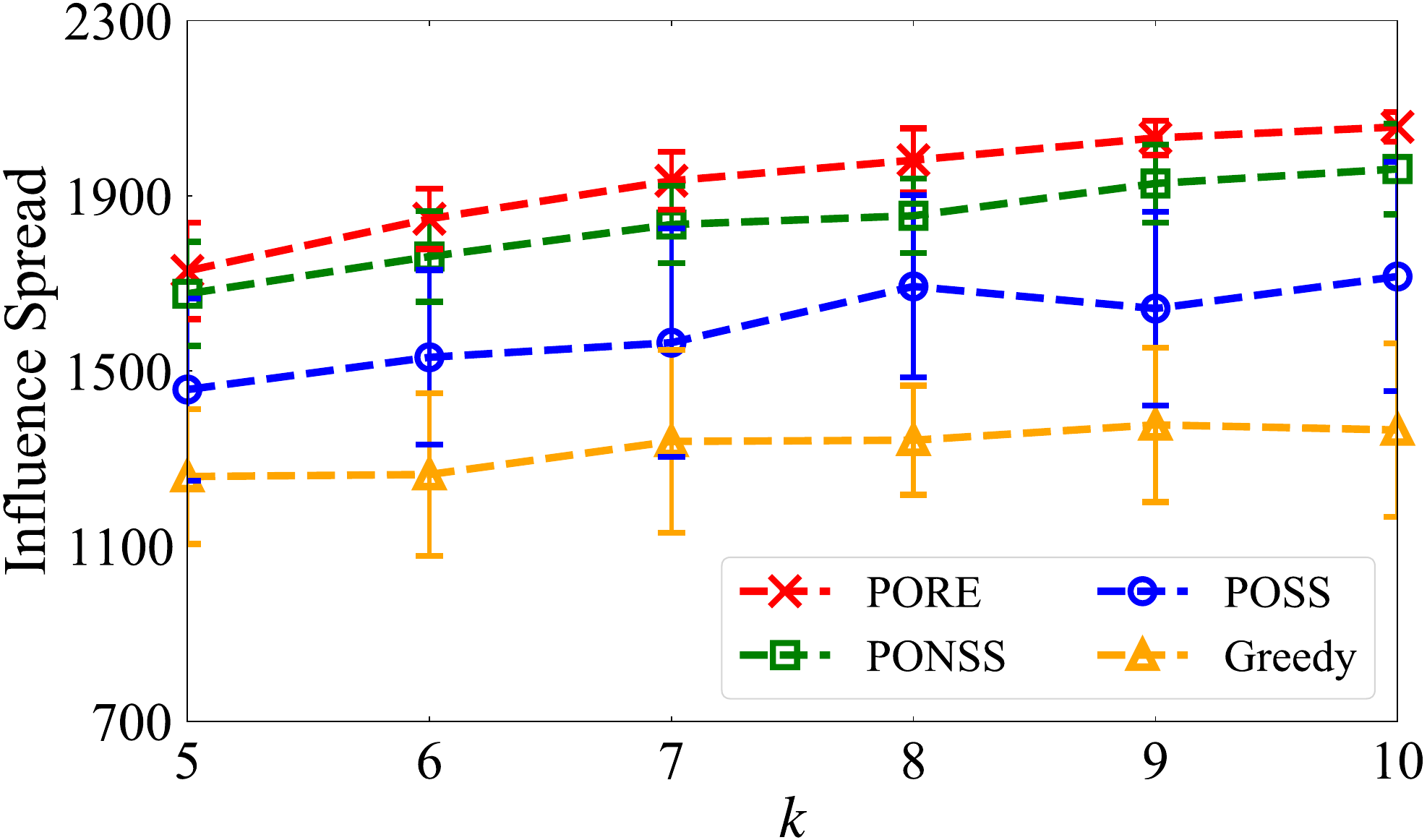} 
      \hspace{1cm}
      \includegraphics[width=0.36\linewidth]{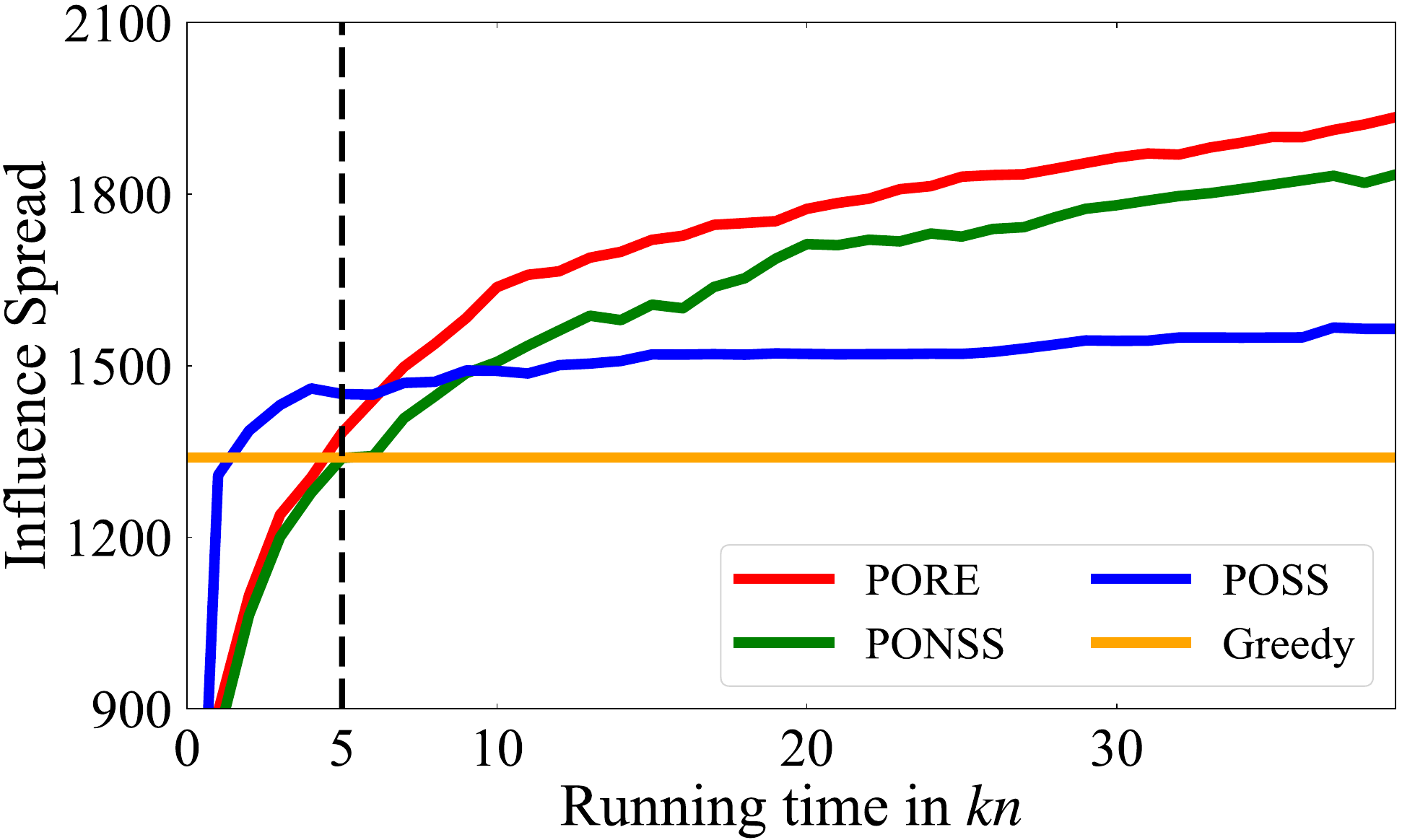}
      \caption{\textit{ego-Facebook} (4,039 \#nodes, 88,234 \#edges)}
      \label{fig:dataset1}
    \end{subfigure}
    \vspace*{0.2cm}
    \begin{subfigure}{\textwidth}
      \centering
      \includegraphics[width=0.36\linewidth]{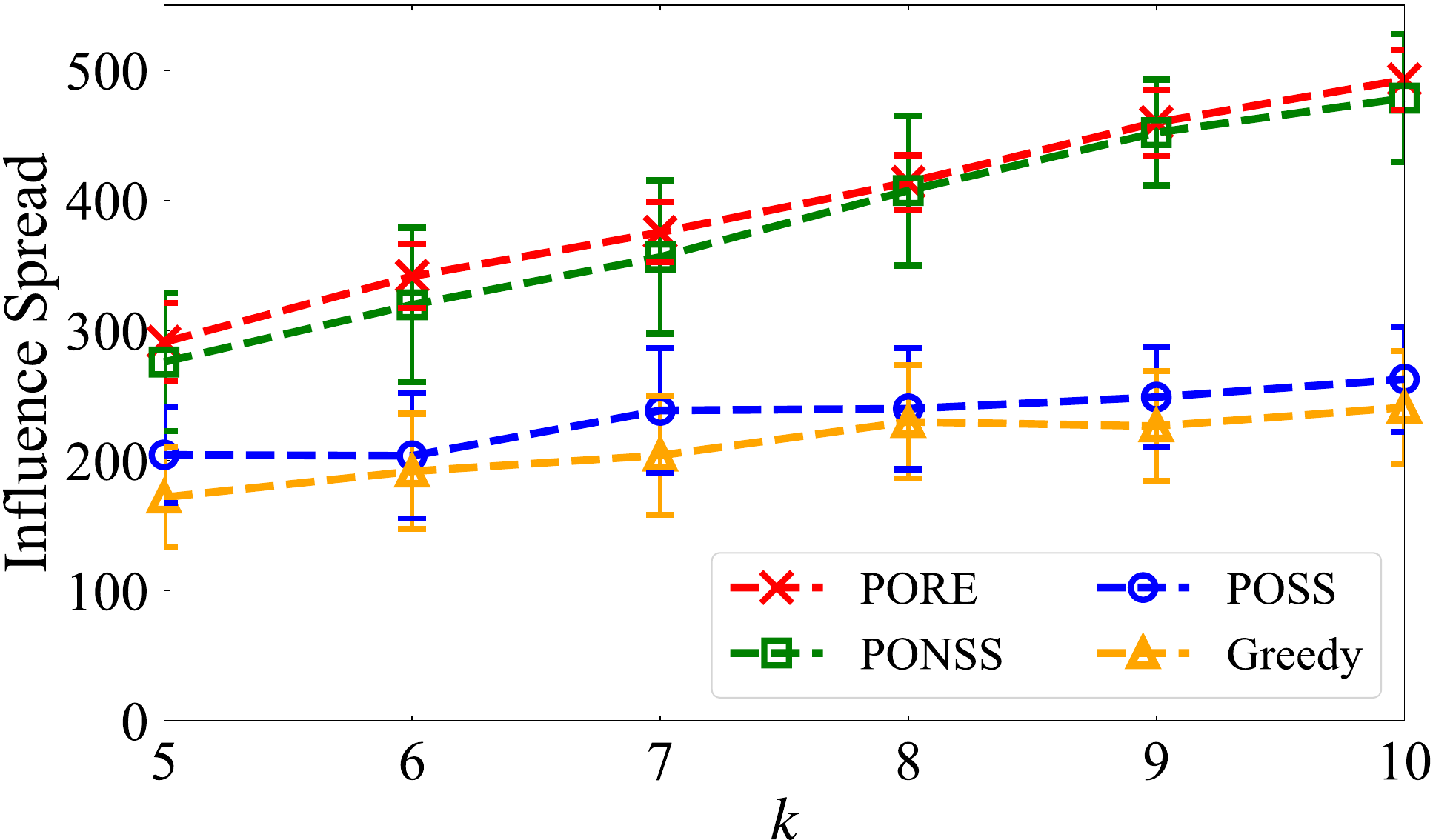}
      \hspace{1cm}
      \includegraphics[width=0.36\linewidth]{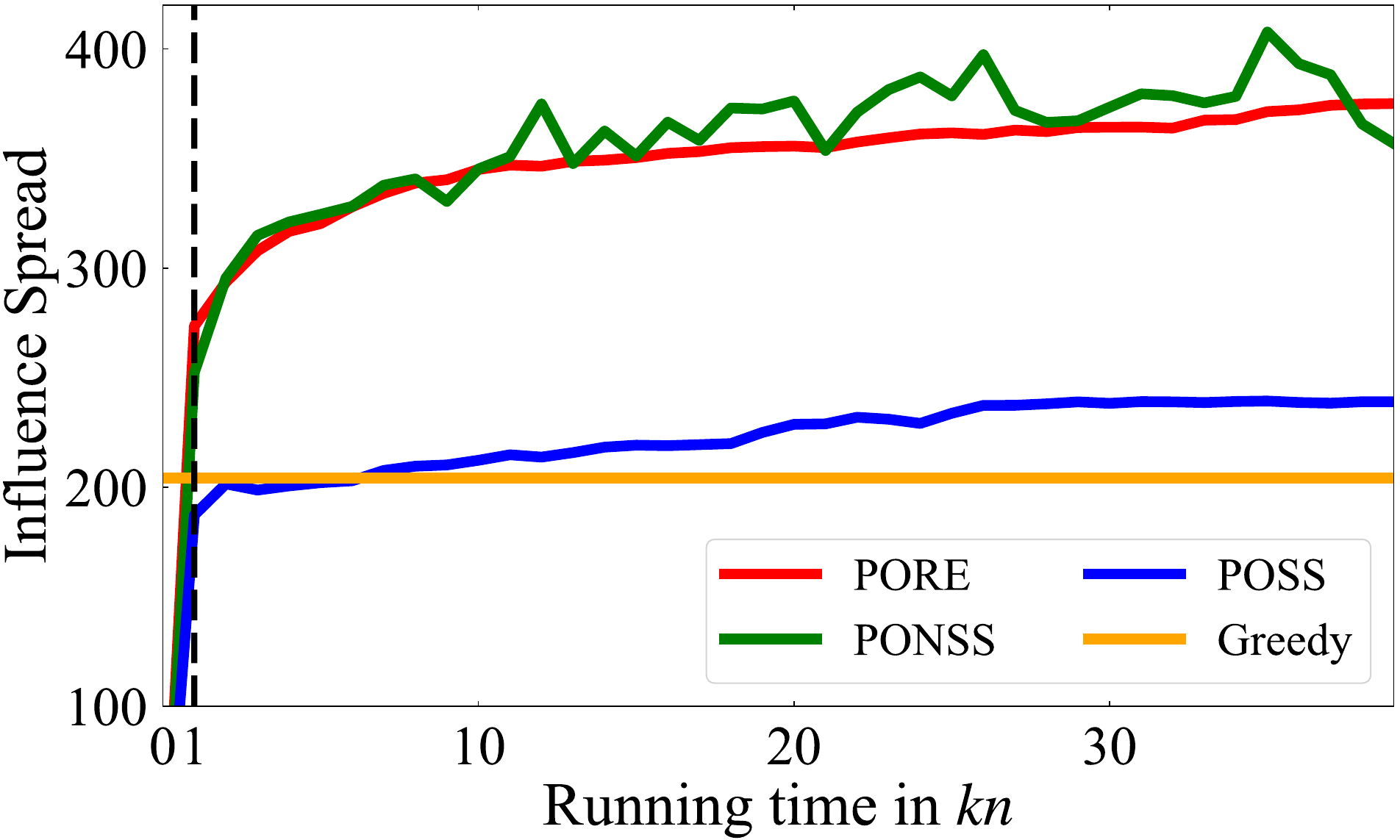}
      \caption{\textit{HepPh} (12006 \#nodes, 236978 \#edges)}
      \label{fig:dataset2}
    \end{subfigure}
    
    \vspace{-5pt}
    
    \caption{Influence maximization (influence spread: the larger the better). The left subfigure on each dataset: influence spread vs budget $k$. The right subfigure on each dataset: influence spread vs running time of PORE, PONSS and POSS for $k=7$.}
    \label{fig:IM}
    \vspace{-5pt}
\end{figure*}
\subsection{Influence Maximization}

We use two real-world datasets: \textit{ego-Facebook} and \textit{HepPh}\footnote{The ego-Facebook dataset, used in the PONSS paper~\cite{qian2017subset}, is available from its open-source code, whereas the other dataset from the same source is now not publicly accessible. Consequently, we use the similarly sized HepPh dataset.}. In our analysis, the propagation probability of one edge from node $u$ to $v$ is estimated by $\frac{weight(u, v)}{indegree(v)}$. This method of estimation is widely recognized and has been extensively applied in previous studies such as~\cite{chen2009efficient,goyal2011simpath}. For the parameters used in the algorithms, we set the parameter $\theta$ in Definition~\ref{def5} to 0.15. We set $k$ from \{5, 6, \ldots, 10\}. To estimate the objective, influence spread, we independently simulate the diffusion process on the network 10 times and take the average as the estimation, which is thus noisy. However, for the final output solutions of the algorithms, we conduct 10,000 independent simulations to ensure a more accurate estimation consistent with~\cite{qian2017subset}. We plot the average results of 30 independent runs in Fig.~\ref{fig:IM}, where the error bars around each line indicate the standard deviation (std).

In Fig.~\ref{fig:IM}, the left subfigure for each dataset illustrates the performance of the algorithms under various cardinalities $k$. We can find that the performance of each algorithm generally improves as $k$ increases. It is reasonable because a larger subset will exhibit larger influence spread. It can be seen that PORE performs the best among all algorithms, showing its superiority under noisy environment. Meanwhile, PORE demonstrates superior stability, as evidenced by its narrower std compared to the other algorithms. Consistent with the work of Qian et al.~\cite{qian2017subset}, the performance of the PONSS algorithm surpasses that of the POSS algorithm, which in turn outperforms the greedy algorithm. By selecting the greedy algorithm as the baseline, we plot the right subfigures of Fig.~\ref{fig:IM}, i.e., the curve of influence spread over running time for PORE, PONSS and POSS with $k=7$. Note that the $x$-axis is in $kn$, the running time required by the greedy algorithm. The greedy algorithm is a fixed-time algorithm, while others (POSS, PONSS and PORE) are anytime algorithms, and can get better performance by using more time (less than $10kn$ in the right subfigures of Fig.~\ref{fig:IM}). It is evident that PORE quickly achieves superior performance on the dataset \textit{ego-Facebook}. Additionally, for the dataset \textit{HepPh} with a more complex and large network, the optimization process of PONSS exhibits instability, while PORE does not.

\subsection{Sparse Regression}
\begin{figure*}
    \begin{subfigure}{\textwidth}
      \centering
      \includegraphics[width=0.36\linewidth]{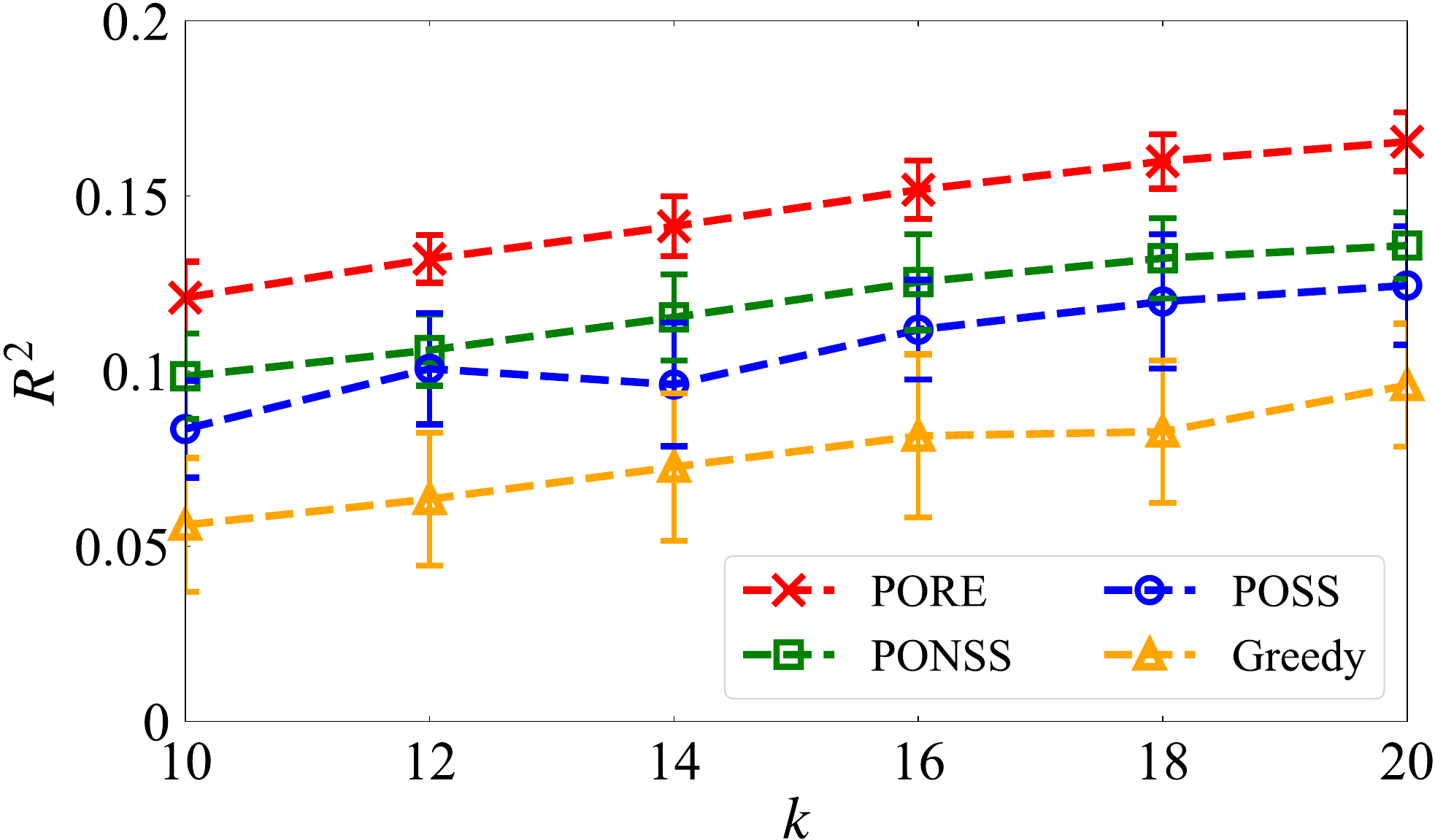} 
      \hspace{1cm}
      \includegraphics[width=0.36\linewidth]{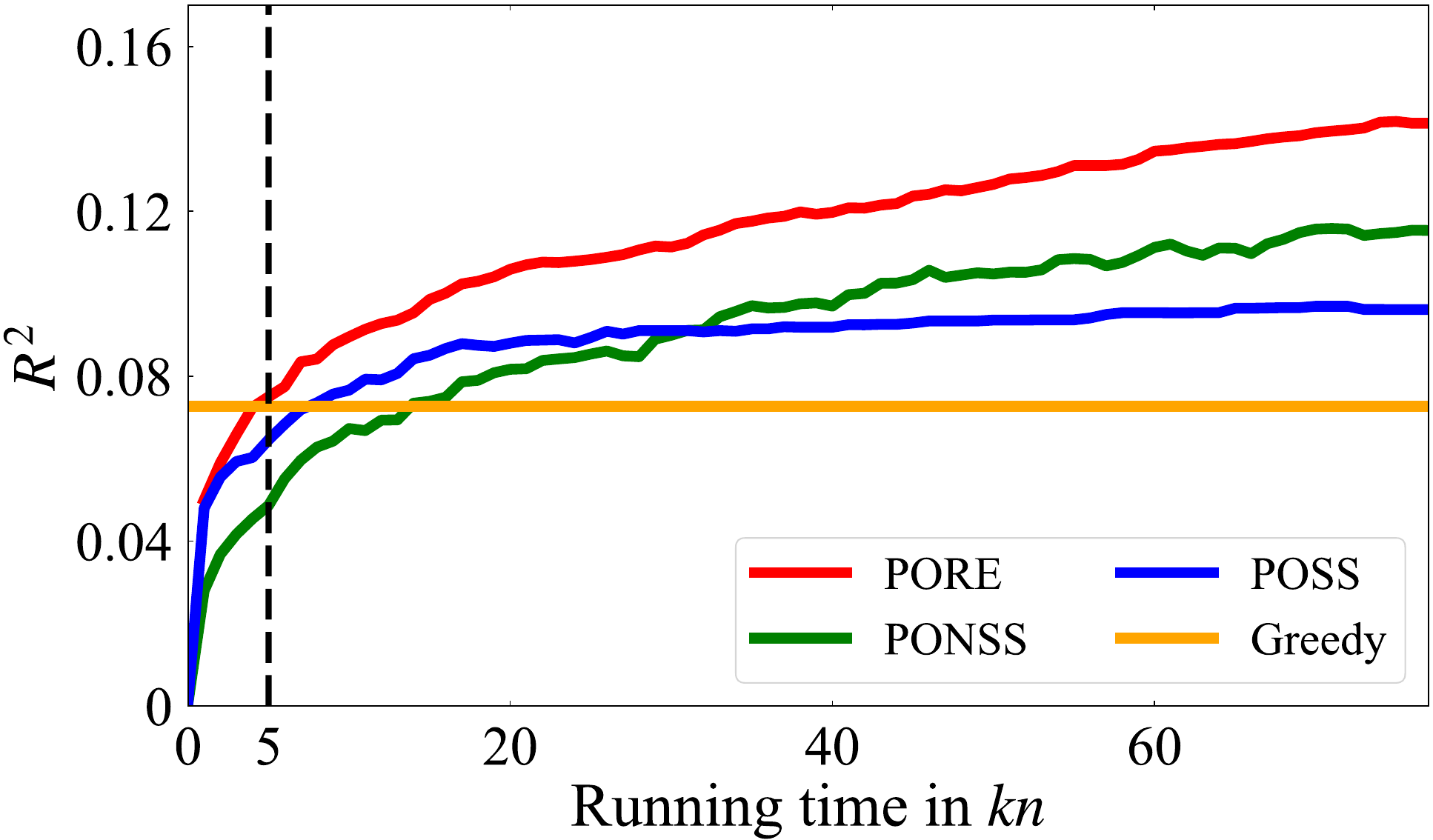}
      \caption{\textit{protein} (24,387 \#instances, 357 \#features)}
      \label{fig:dataset3}
    \end{subfigure}
    
    \vspace*{0.2cm}

    \begin{subfigure}{\textwidth}
      \centering
      \includegraphics[width=0.36\linewidth]{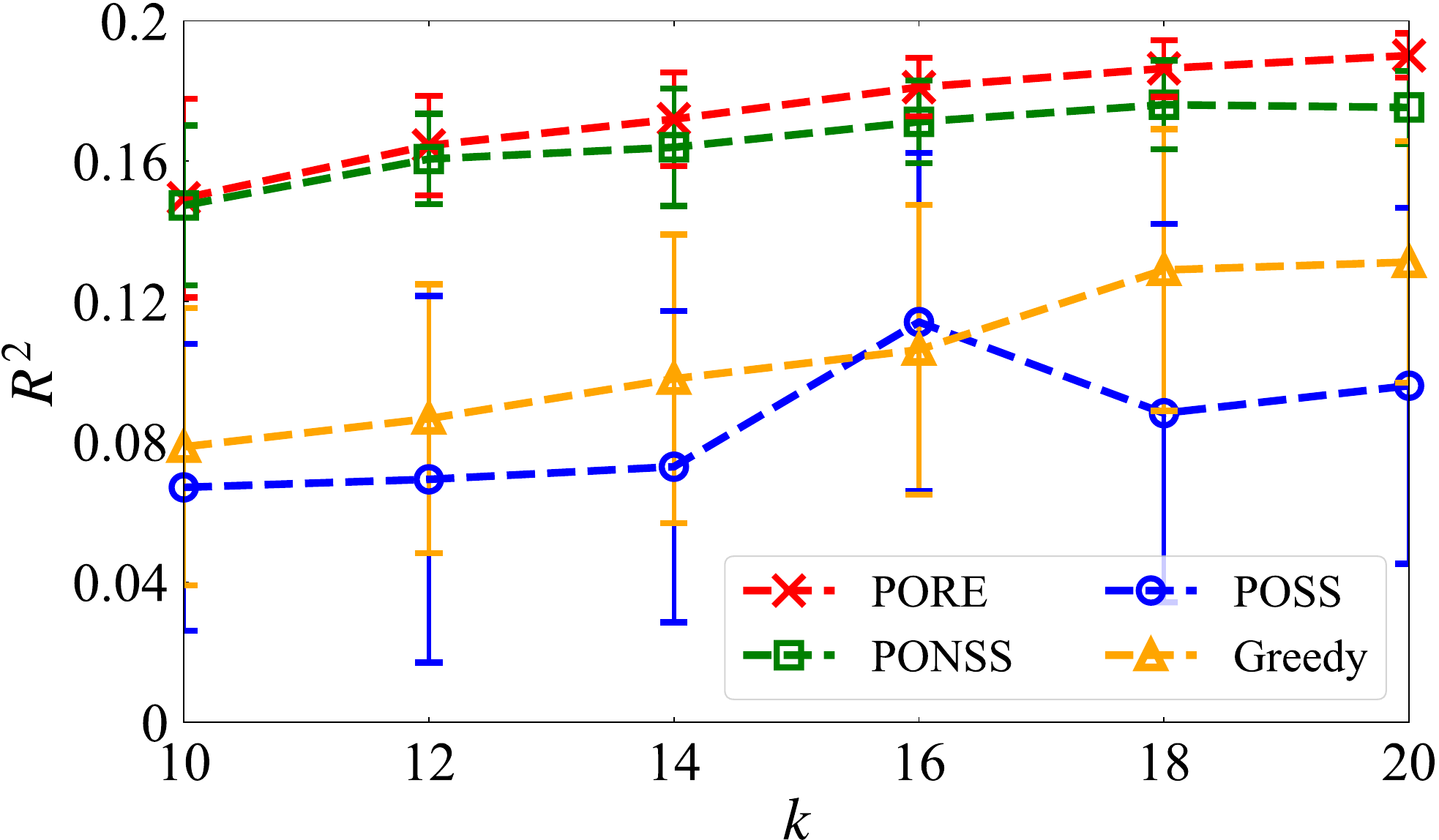}
      \hspace{1cm}
      \includegraphics[width=0.36\linewidth]{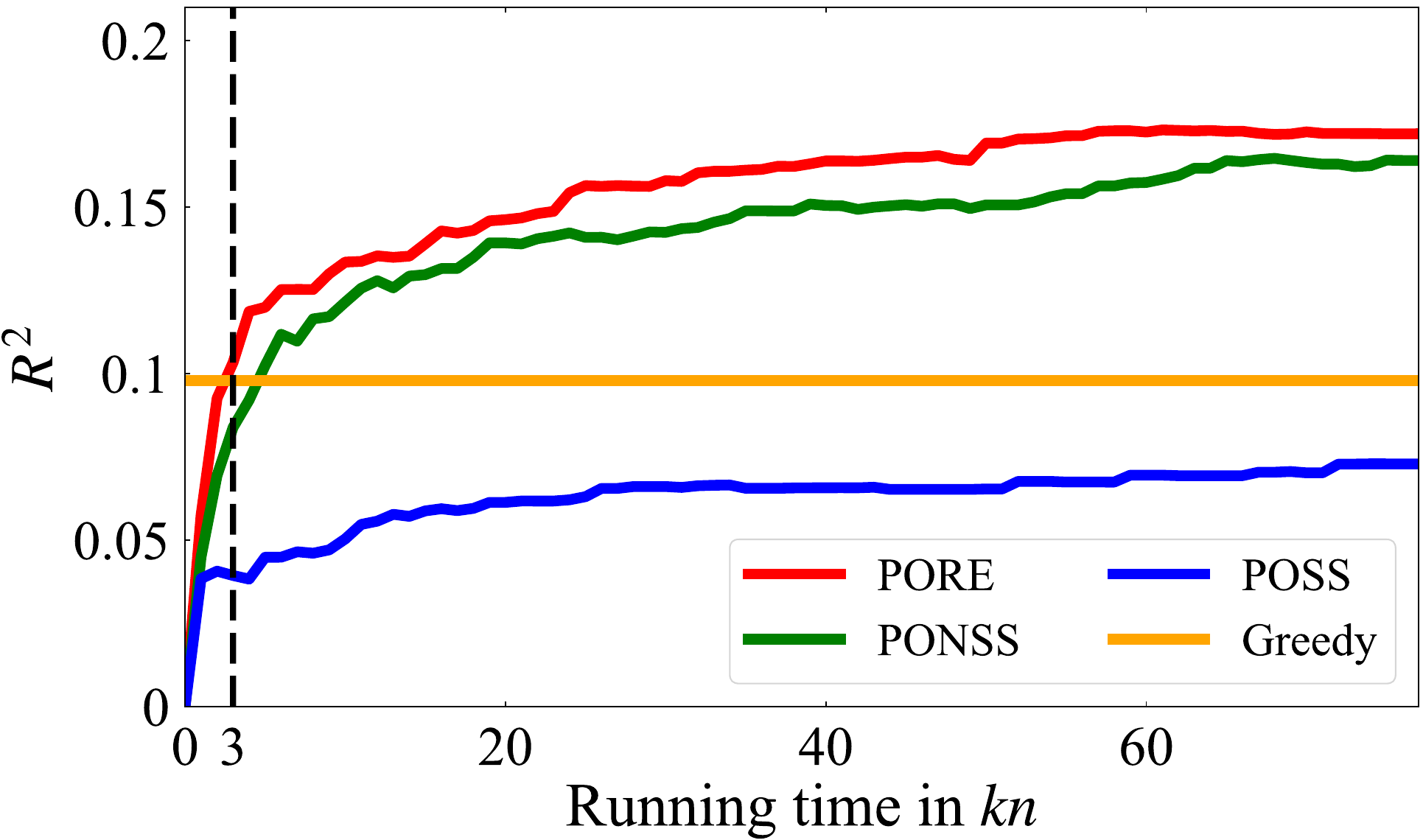}
      \caption{\textit{YearPredictionMSD} (515,345 \#instances, 90 \#features)}
      \label{fig:dataset4}
    \end{subfigure}

    \vspace{-5pt}
  
    \caption{Sparse regression ($R^2$: the larger the better). The left subfigure on each dataset: $R^2$ vs budget $k$. The right subfigure on each dataset: $R^2$ vs
    running time of PORE, PONSS and POSS for $k = 14$.}
    \label{fig:SR}
    \vspace{-5pt}
\end{figure*}
We use the same datasets in~\cite{qian2017subset}, namely \textit{protein} and \textit{YearPredictionMSD}. The $k$ varies from $\{10, 12, \dots, 20\}$ and $\theta$ is set to 0.05. For estimating $R^2$ in the optimization process, we use a random sample of 1000 instances, leading to a noisy evaluation. But for the final output solutions, we use the whole dataset for accurate estimation. The average results are plotted in Fig.~\ref{fig:SR}. We can observe from the left subfigures of Fig.~\ref{fig:SR} that PORE consistently achieves the best results on all cases with various $k$. The right subfigures of Fig.~\ref{fig:SR} demonstrate that PORE can quickly achieve superior performance within $5kn$ time.

\subsection{Ablation Study of Robust Evaluation}
\begin{figure*}
    \centering
    \begin{subfigure}{.36\textwidth}
      \centering
      \includegraphics[width=\linewidth]{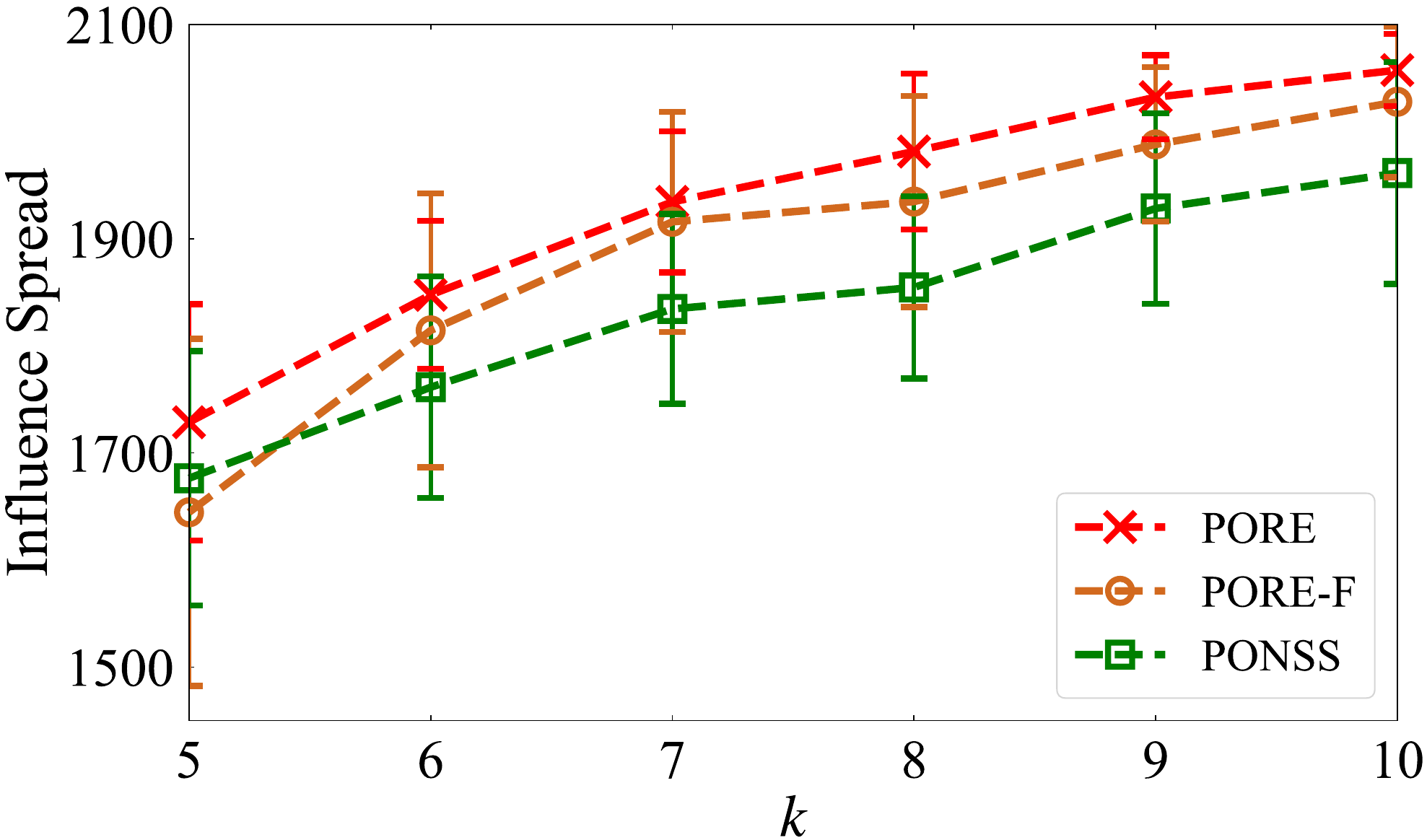}
      \caption{\textit{ego-Facebook}}
      \label{fig:Ablsub1}
    \end{subfigure}%
    \hspace{1cm}
    \begin{subfigure}{.36\textwidth}
      \centering
      \includegraphics[width=\linewidth]{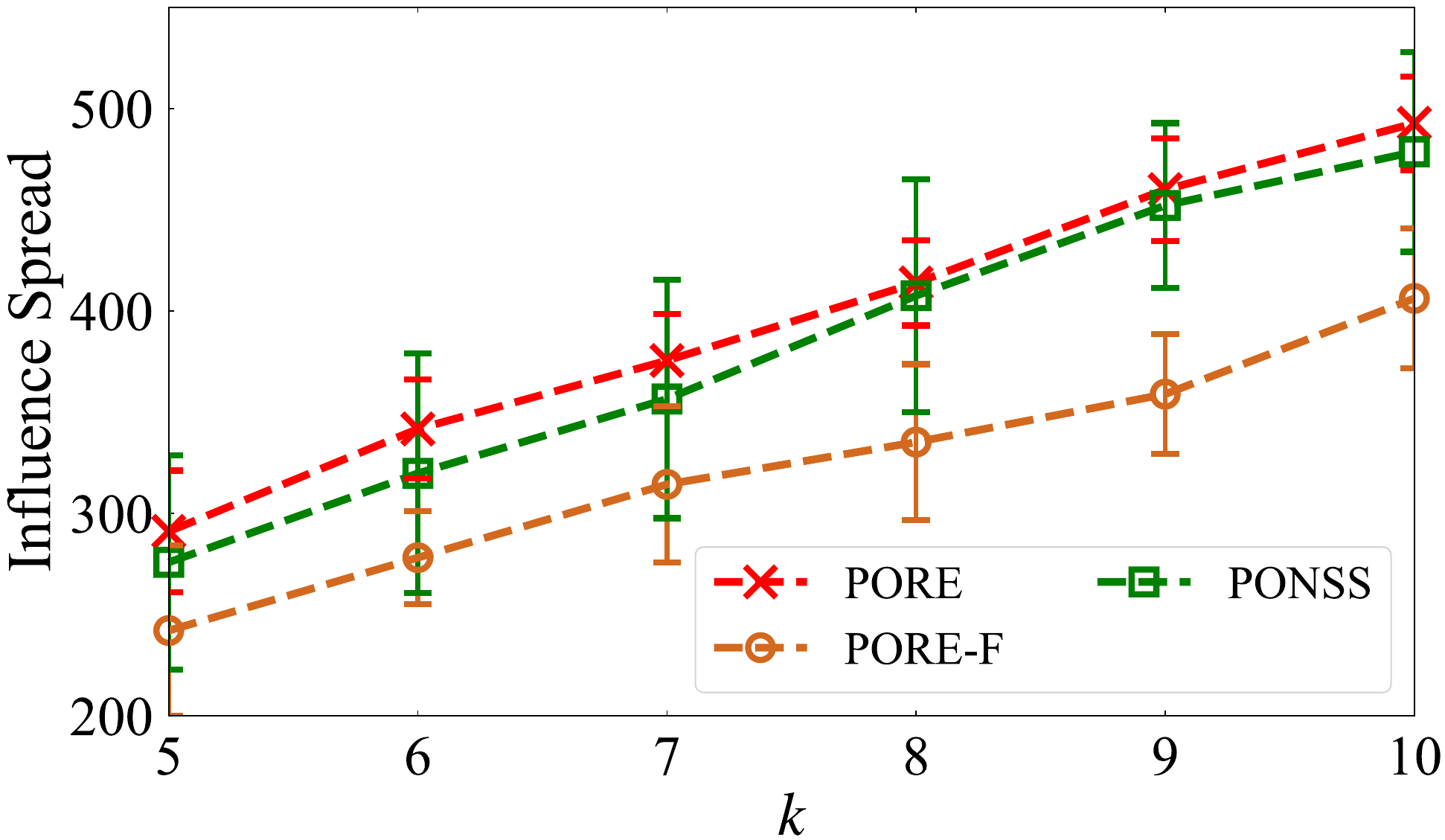}
      \caption{\textit{HepPh}}
      \label{fig:Ablsub2}
    \end{subfigure}
    
    \begin{subfigure}{.36\textwidth}
      \centering
      \includegraphics[width=\linewidth]{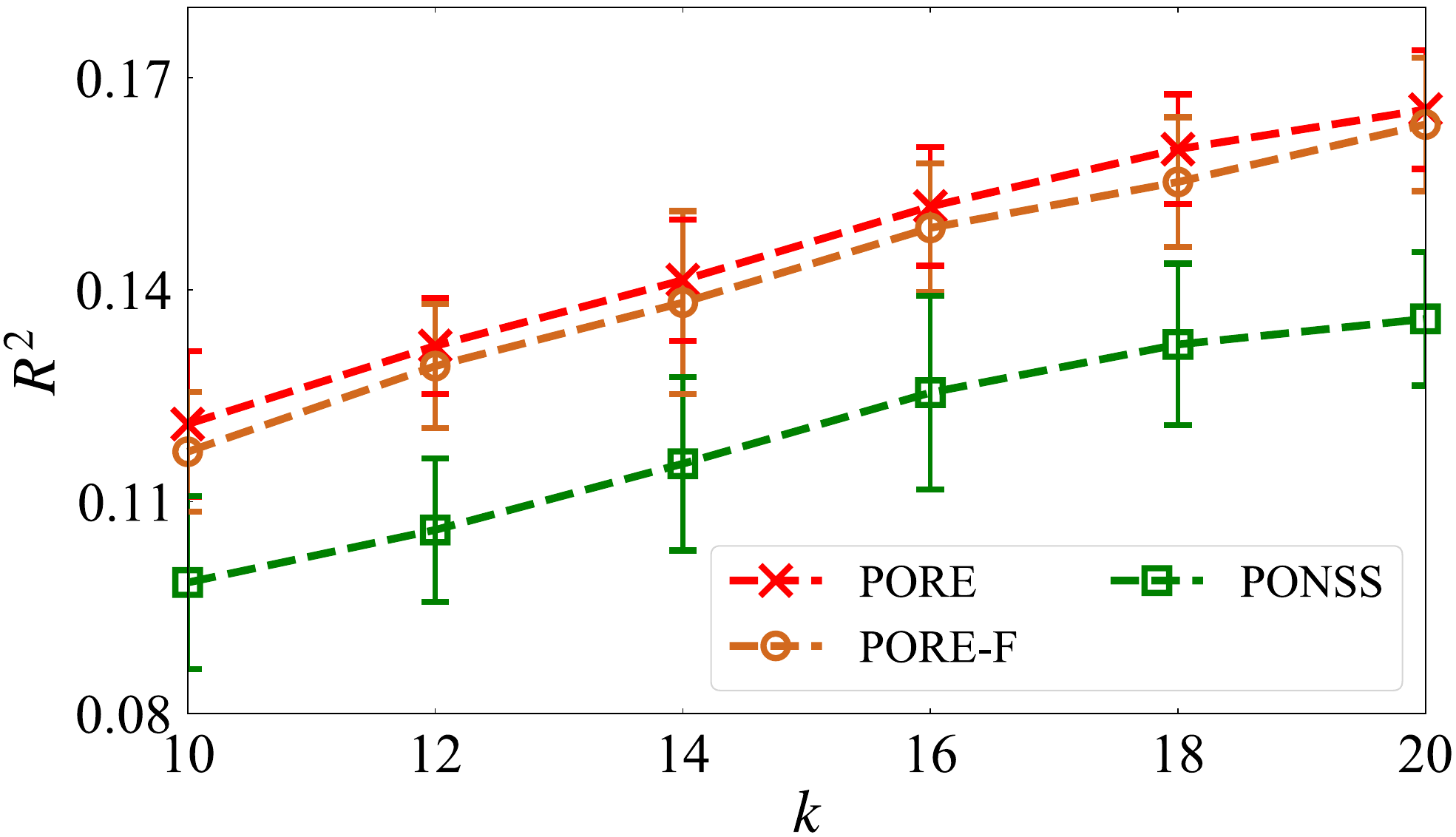}
      \caption{\textit{protein}}
      \label{fig:Ablsub3}
    \end{subfigure}%
    \hspace{1cm}
    \begin{subfigure}{.36\textwidth}
      \centering
      \includegraphics[width=\linewidth]{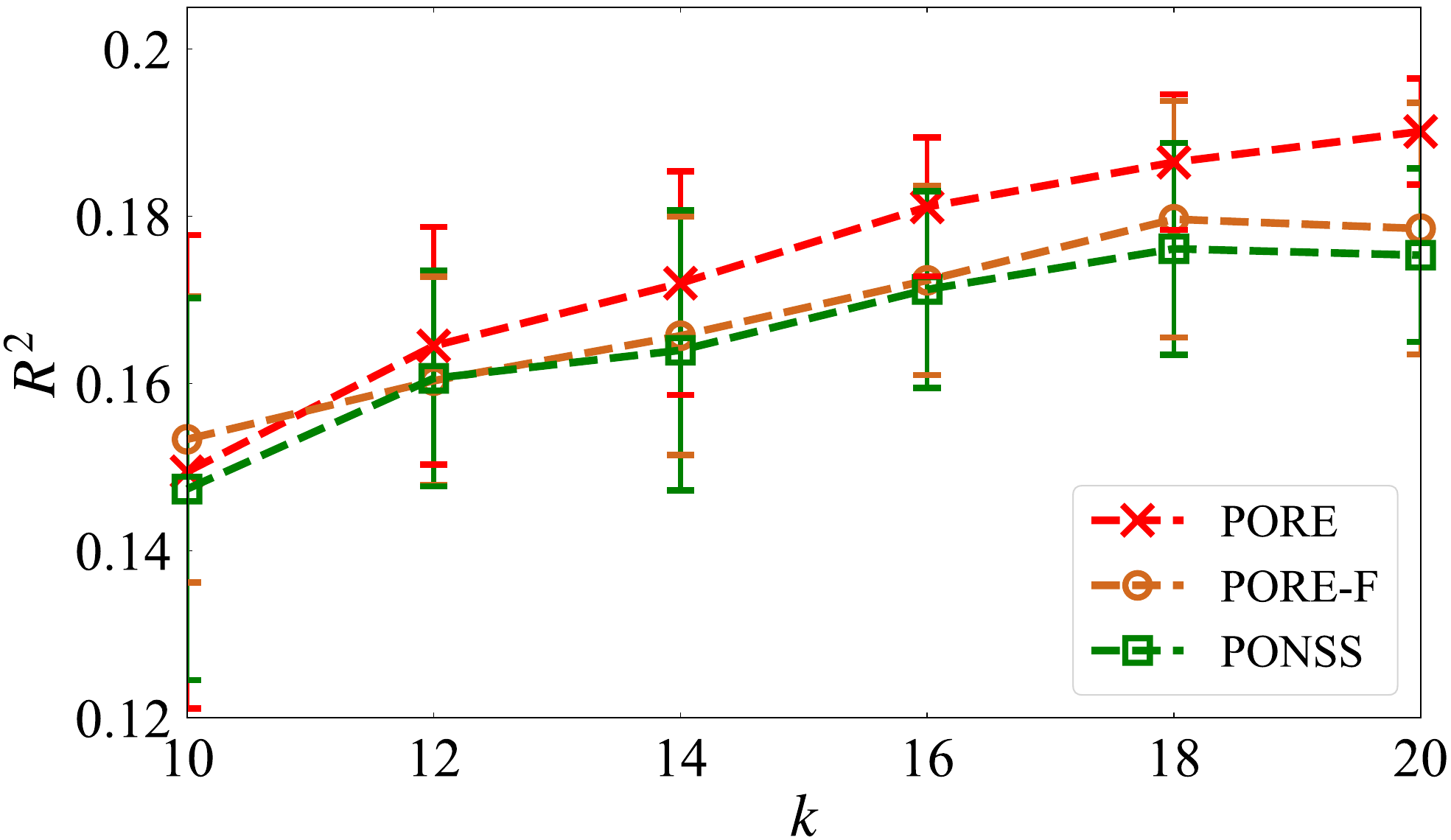}
      \caption{\textit{YearPredictionMSD}}
      \label{fig:Ablsub4}
    \end{subfigure}
    
    \caption{The ablation experiments on the robust evaluation, where PORE-F denotes PORE without robust evaluation.}
    \label{fig:Abl}
\end{figure*}

To verify the effectiveness of robust evaluation, we run the PORE-F algorithm modified from PORE by setting $f_1(\bm{x})=F(\bm{x})$ rather than the robust evaluation in Eq.~(\ref{eq-bi-PORE}). The experimental results on both two applications are shown in Fig.~\ref{fig:Abl}. It can be seen that PORE-F performs worse than PORE, and is sometimes even the worst performer among the three algorithms in Fig.~\ref{fig:Abl}(b), which implies the effectiveness of robust evaluation.

\subsection{Comparison under Different Noise Intensities} 
\begin{figure*}
    \centering
    \begin{subfigure}{.36\textwidth}
      \centering
      \includegraphics[width=\linewidth]{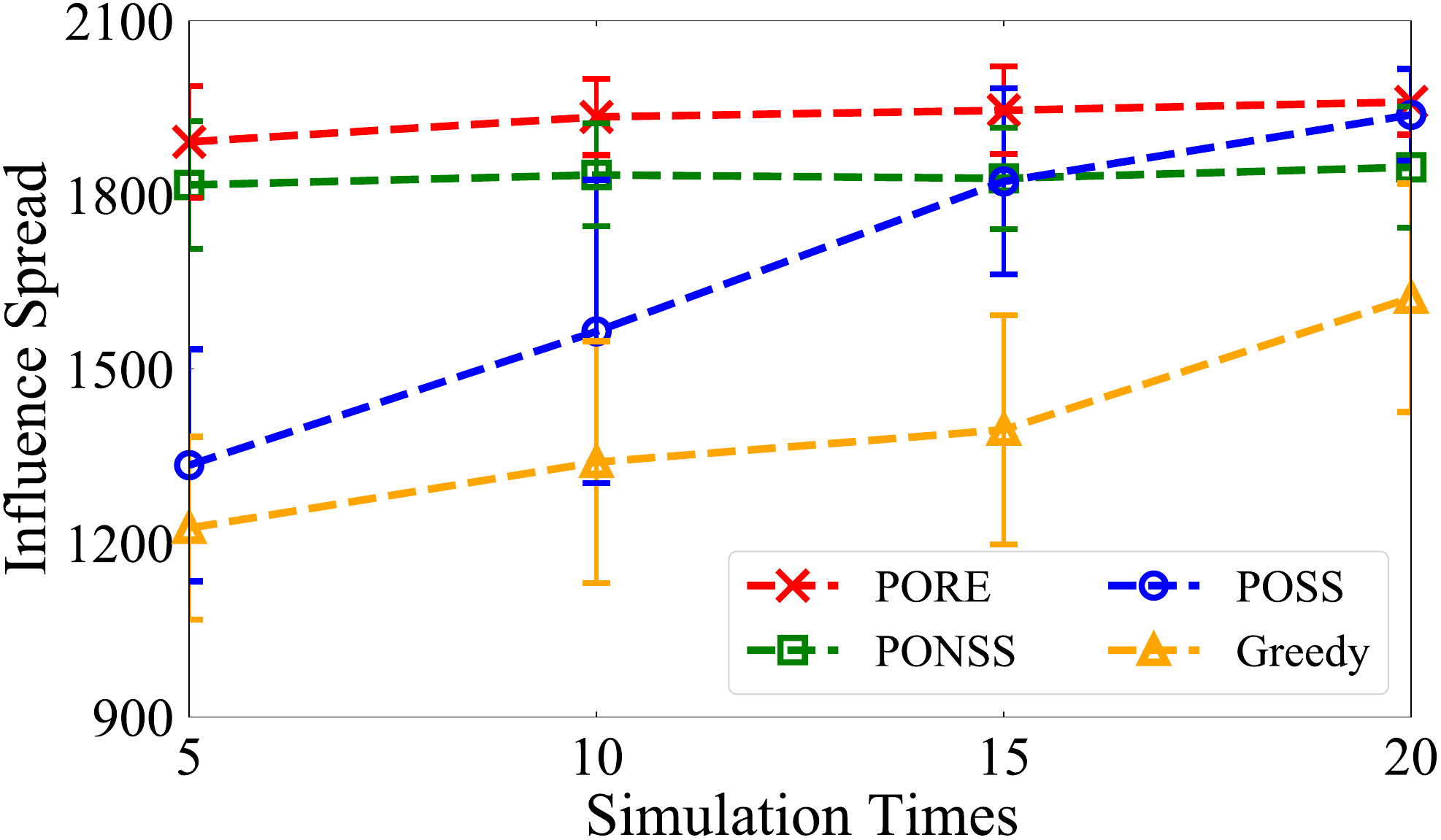}
      \caption{\textit{ego-Facebook}}
      \label{fig:noise1}
    \end{subfigure}%
    \hspace{1cm}
    \begin{subfigure}{.36\textwidth}
      \centering
      \includegraphics[width=\linewidth]{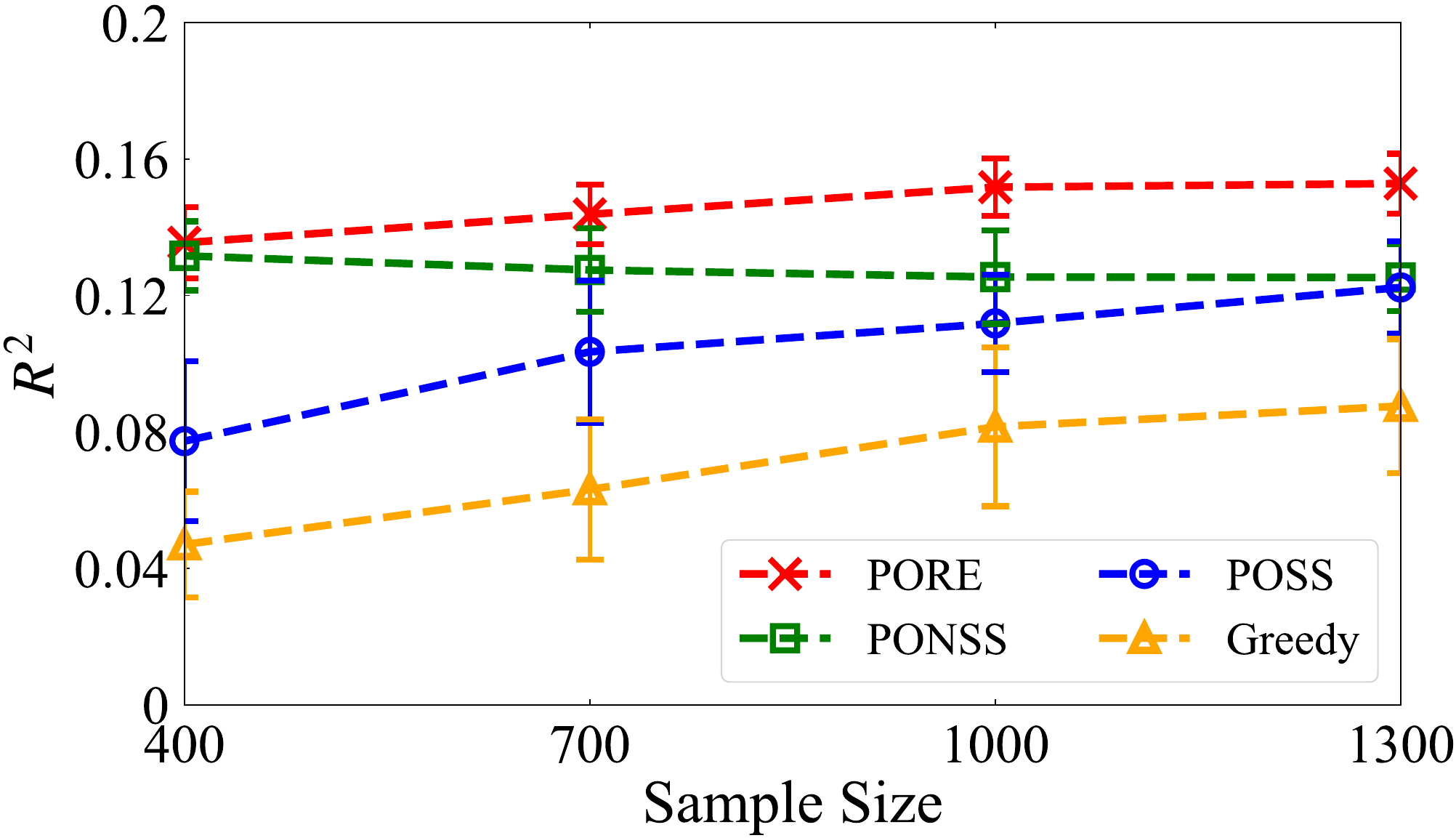}
      \caption{\textit{protein}}
      \label{fig:noise2}
    \end{subfigure}

    \vspace{-5pt}
    
    \caption{Performance of the algorithms under different noise intensities. The left subfigure on \textit{ego-Facebook} displays the influence spread versus simulation times for $k=7$. The right subfigure on \textit{protein} displays $R^2$ versus sample size for $k=16$.}
    \label{fig:noise}

    \vspace{-5pt}
    
\end{figure*}
We compare the performance of PORE, PONSS, POSS, and the greedy algorithm under different noise intensities. For influence maximization, the average of simulation runs to estimate the influence spread leads to noise. The more simulation runs, the smaller the noise. Thus, we run experiments by varying simulation times from $\{5, 10, 15, 20\}$ to compare the performance of each algorithm on \textit{ego-Facebook} with $k=7$. For sparse regression, a sample of instances instead of all the instances used to calculate $R^2$ brings noise. The more sampled instances, the smaller the noise. Thus, we run experiments by setting different sample sizes from $\{400, 700, 1000, 1300\}$ on \textit{protein} with $k=16$. As expected, Fig.~\ref{fig:noise} shows that the performance of algorithms designed for noise-free environments (i.e., POSS and the greedy algorithm) improves with the evaluation accuracy. That is, as the noise level decreases, POSS and the greedy algorithm perform better. The noise-resistant algorithms (i.e., PONSS and PORE) perform relatively stable under different noise intensities, and PORE always performs the best.

\subsection{Influence of Different Settings of $\theta$}
\begin{figure}[h]
    \centering
    \includegraphics[width=0.35\textwidth]{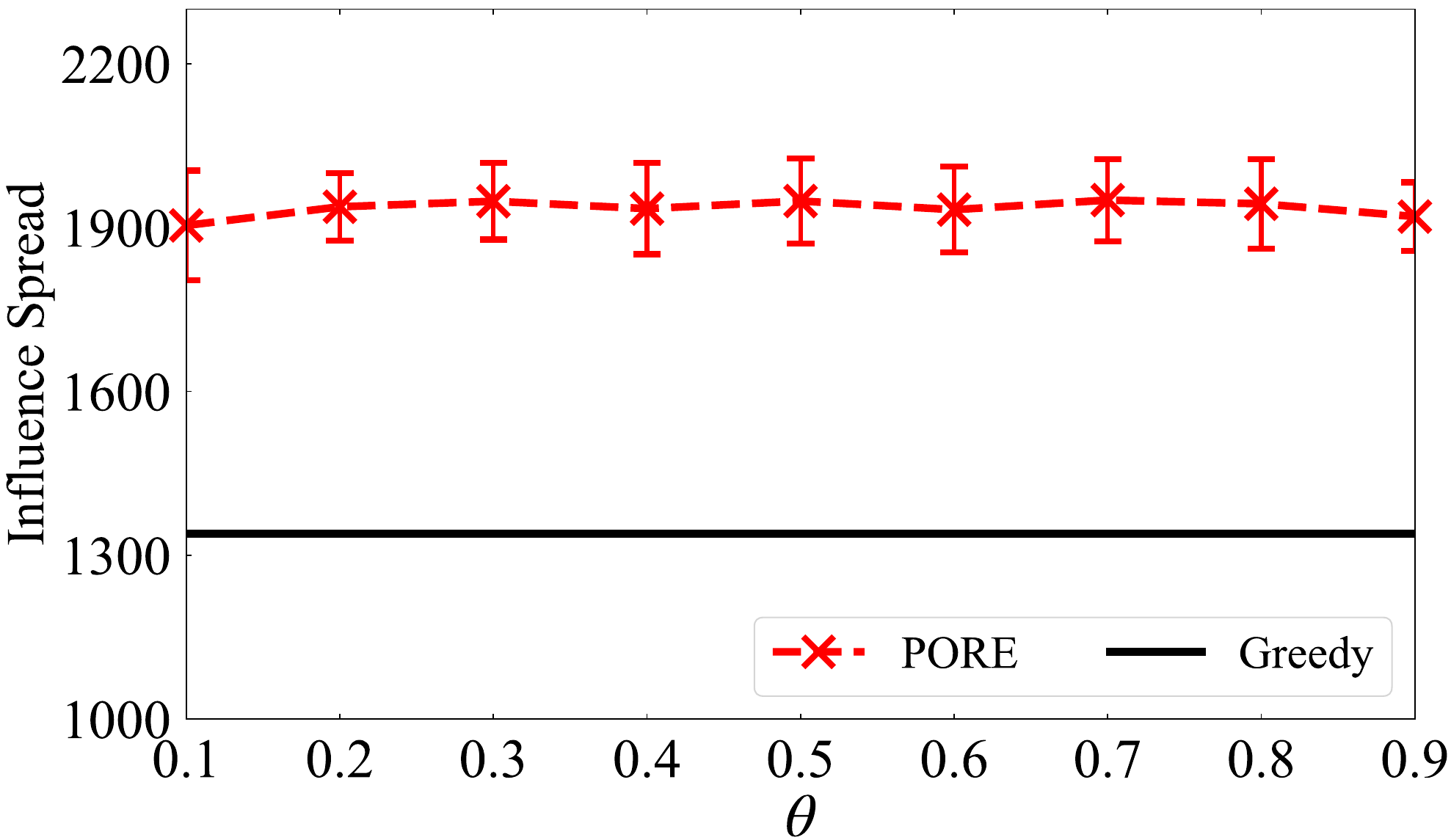}
    \caption{Performance of PORE with different $\theta$ values on the application of influence maximization, where the dataset is \textit{ego-Facebook} and the budget $k=7$.}
    \label{fig:theta}
\end{figure}

The value of $\theta$ in $\theta$-domination used in PORE serves a hyperparameter. To examine the influence of $\theta$ on the performance of the PORE algorithm, we vary $\theta$ from 0.1 to 0.9 in increments of 0.1 and run PORE on the \textit{ego-Facebook} dataset with $k = 7$. As depicted in Fig.~\ref{fig:theta}, it can be observed that the performance of the PORE algorithm is not sensitive to the value of $\theta$.

\subsection{Summary of Results}
To conclude our investigations, the results consistently demonstrate that PORE is superior to the classical greedy algorithm, POSS, and PONSS in terms of both solution quality and stability. By aggregating evaluations from structural neighbors, PORE effectively filters out stochastic outliers that often mislead POSS and the greedy algorithm. Unlike the computational overhead incurred by PONSS through repetitive re-evaluations, PORE utilizes its evaluation budget more strategically, achieving faster convergence and higher $R^2$ or influence spread values within the same total evaluation limit. Furthermore, the ablation studies and noise analysis underscore the efficacy of our robust evaluation function $f_1$, which maintains high performance even as noise levels increase. The insensitivity of PORE to the hyperparameter $\theta$ suggests its practicality for real-world scenarios where precise parameter tuning may be difficult. Collectively, these observations confirm that PORE provides a more reliable and versatile framework by prioritizing structural robustness during the search, ensuring steady progress toward high-quality solutions even amidst significant environmental uncertainty.

\section{Conclusion}

In this paper, we propose a new algorithm based on Pareto optimization with robust evaluation for noisy subset selection, named PORE. The robust evaluation method assesses the quality of a solution by calculating the average value of a noisy function for its neighboring solutions, each of which has exactly one element less than the original solution, which can more accurately determine the quality of solutions under noisy environments. Experiments on influence maximization and sparse regression demonstrate the superior performance of PORE over previous algorithms, i.e., the classic greedy algorithm, POSS and PONSS. In the future, it will be interesting to theoretically analyze the approximation performance of PORE.

\begin{acks}
This work was supported by the Science and Technology Project of the State Grid Corporation of China (Research on Electromagnetic Transient Simulation Technology Based on AI for Science, 52170025002P-465-FGS). Chao Qian is the corresponding author.
\end{acks}

\vspace{15pt}

\bibliographystyle{ACM-Reference-Format}
\bibliography{PORE}

\appendix

\end{document}